\documentclass[10pt,twocolumn,letterpaper]{article}

\usepackage{cvpr}              % To produce the CAMERA-READY version
\definecolor{cvprblue}{rgb}{0.21,0.49,0.74}
\usepackage[pagebackref,breaklinks,colorlinks,allcolors=cvprblue]{hyperref}
\usepackage{multirow}
\usepackage{arydshln}
\hypersetup{colorlinks=true}
\usepackage{marvosym}
\def\paperID{8769} % *** Enter the Paper ID here
\def\confName{CVPR}
\def\confYear{2025}

\newcommand{\paperName}{PhysFlow}
\title{Unleashing the Potential of Multi-modal Foundation Models and Video Diffusion for 4D Dynamic Physical Scene Simulation}

\author{
Zhuoman Liu\textsuperscript{1} \quad
Weicai Ye\textsuperscript{2,\textrm{\Letter},}\thanks{: Project Lead, \textrm{\Letter}: Corresponding Author} \quad
Yan Luximon\textsuperscript{1,\textrm{\Letter}} \quad
Pengfei Wan\textsuperscript{2} \quad
Di Zhang\textsuperscript{2}\\
\textsuperscript{1}The Hong Kong Polytechnic University \quad 
\textsuperscript{2}Kuaishou Technology\\
\small{\texttt{maikeyeweicai@gmail.com}}
\quad
\small{\texttt{yan.luximon@polyu.edu.hk}}
\quad
\small{\texttt{zhuo-man.liu@connect.polyu.hk}}
}

\begin{document}
\maketitle
\begin{abstract}
Realistic simulation of dynamic scenes requires accurately capturing diverse material properties and modeling complex object interactions grounded in physical principles. However, existing methods are constrained to basic material types with limited predictable parameters, making them insufficient to represent the complexity of real-world materials. We introduce \textbf{\paperName{}}, a novel approach that leverages multi-modal foundation models and video diffusion to achieve enhanced 4D dynamic scene simulation. Our method utilizes multi-modal models to identify material types and initialize material parameters through image queries, while simultaneously inferring 3D Gaussian splats for detailed scene representation. We further refine these material parameters using video diffusion with a differentiable Material Point Method (MPM) and optical flow guidance rather than render loss or Score Distillation Sampling (SDS) loss. This integrated framework enables accurate prediction and realistic simulation of dynamic interactions in real-world scenarios, advancing both accuracy and flexibility in physics-based simulations. Our code and data are available at \href{https://zhuomanliu.github.io/PhysFlow}{https://zhuomanliu.github.io/PhysFlow}
\end{abstract}    
\vspace{-1em}
\section{Introduction}
\label{sec:intro}

The realistic simulation of dynamic scenes in computer vision and graphics is critical for applications such as robotics manipulation~\cite{xu2020learning,shi2023robocook} and video generation~\cite{blattmann2023stable,yang2024cogvideox,dreamscene4d}. However, existing video generation methods often produce visually plausible results that lack physical realism, leading to inaccuracies in object motion and deformation. To enhance simulation realism, it is essential to automatically capture the diverse material properties of objects and model physical interactions under varying forces and conditions.

Among existing methods, PAC-NeRF~\cite{li2023pac} and GIC~\cite{cai2024gic} aim to learn material parameters by analyzing the deformation of objects in multi-view dynamic videos. PAC-NeRF integrates neural radiance fields (NeRF)~\cite{mildenhall2021nerf} with a differentiable Material Point Method (MPM)~\cite{jiang2015mpm} to facilitate the simulation of various material types and identify material properties. However, it is constrained to simpler objects and textures, leading to lower rendering fidelity that limits its effectiveness in high-quality dynamic scene simulations. GIC uses 3D Gaussian Splatting~\cite{kerbl3Dgaussians} to capture explicit shapes and relies on continuum mechanics to infer implicit shapes, assisting in estimating physical properties. Nonetheless, PAC-NeRF and GIC require multi-view images of deforming objects with known camera poses and prior knowledge of material types, which restricts its applicability in arbitrary dynamic scene videos.

PhysGaussian~\cite{xie2024physgaussian} utilizes 3D Gaussian Splatting with differentiable MPM to enhance rendering fidelity, but requires manual initialization of material properties, limiting its adaptability to complex dynamic scenarios. Recent methods~\cite{zhang2024physdreamer,huang2024dreamphysics,liu2024physics3d} use video diffusion for motion guidance to estimate material properties. However, these methods are restricted to elastic material and minor object movements. 
Besides, these methods utilize render loss~\cite{zhang2024physdreamer} or Score Distillation Sampling (SDS)~\cite{poole2022dreamfusion} to estimate material parameters, which presents challenges: render loss is constrained to small motions due to instability and noise in generated videos, while SDS loss demands high computation memory. In contrast, we investigate that optical flow guidance is both memory-efficient and better suited for capturing large and complex motions, enabling more accurate material parameter optimization (see Tab.~\ref{tab:comp_syn} and visual comparisons in the Appendix).

PhysGen~\cite{liu2025physgen} introduces physics reasoning using large pre-trained visual foundation models, eliminating the need for manual parameter initialization. However, it is restricted to single input images and simulates objects at a fixed depth and cannot perform 4D reconstruction or dynamic simulations, limiting its flexibility in simulating complex scenes.

\begin{table*}[t]%\tabcolsep=0.01cm
\centering
\resizebox{1.\linewidth}{!}{
    \begin{tabular}{lcccccc}
        \toprule
         % \textbf{Method} & Various Material Types & w/o Manual Parameter Initialization & \multicolumn{3}{c}{Input Type} & 4D Reconstruction \\
         % & & & Multi-view Images with Poses & Dynamic Video & Single Image & \\
         % \hline
         \textbf{Method} & \textbf{Various Material Types} & \textbf{w/o Manual Init.} & \multicolumn{3}{c}{\textbf{Input Type}} & \textbf{4D Reconstruction} \\
         \cmidrule(lr){4-6}
         & & & \textbf{Multi-view} & \textbf{Dynamic Video} & \textbf{Single Image} & \\
         \midrule
         PAC-NeRF~\cite{li2023pac} & \checkmark (5 types) & & & \checkmark & & \checkmark \\
         GIC~\cite{cai2024gic} & \checkmark (5 types) & & & \checkmark & & \checkmark \\
         \hdashline
         PhysGaussian~\cite{xie2024physgaussian} & \checkmark (6 types) & & \checkmark & & & \checkmark \\
         PhysDreamer~\cite{zhang2024physdreamer} & & & \checkmark & & & \checkmark \\
         Physics3D~\cite{liu2024physics3d} & & & \checkmark & & & \checkmark \\
         \hdashline
         PhysGen~\cite{liu2025physgen} & \checkmark & \checkmark & & & \checkmark & \\
         \hdashline
         \textbf{Ours} & \checkmark (7 types) & \checkmark & \checkmark & \checkmark & \checkmark & \checkmark \\
         \bottomrule
    \end{tabular}
    }
    \vspace{-0.3cm}
    \caption{Comparison of methods by material types, manual parameter initialization, input flexibility, and 4D reconstruction capabilities.}
    \vspace{-0.5cm}
    \label{tab:meth}
\end{table*}

\begin{figure*}[t]
  \centering
   \includegraphics[width=1.\linewidth]{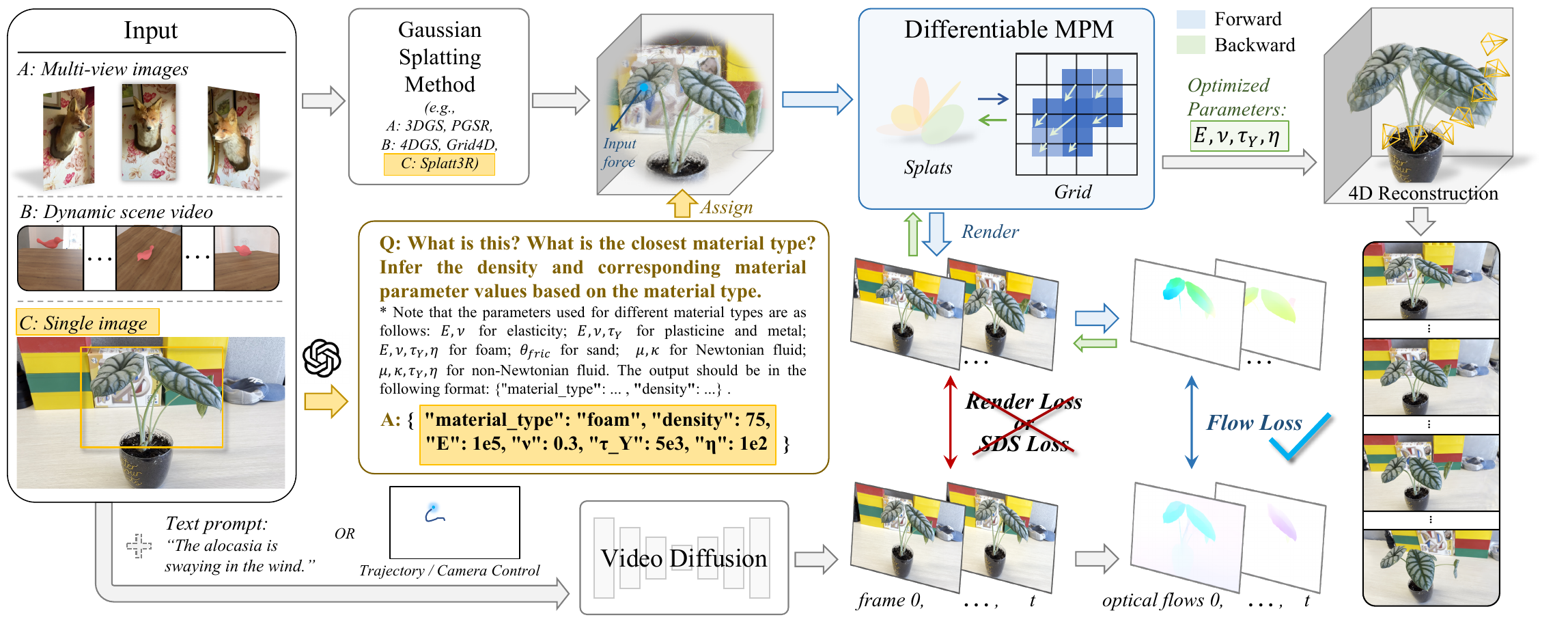}
   \vspace{-0.8cm}
   \caption{Overview of our proposed pipeline for 4D dynamic physical scene simulation. The process begins with 3D scene reconstruction using Gaussian splatting methods for different input types (multi-view images, dynamic video, and single image). Initial material properties are inferred through multi-modal foundation models and assigned to the reconstructed scene. The material parameters are optimized using optical flow guidance within a video diffusion framework, integrated with a differentiable MPM to ensure physically realistic simulation.}
   \label{fig:pipeline}
   \vspace{-0.4cm}
\end{figure*}

To address these challenges, we propose \textbf{\paperName{}}, a novel approach that unleashes the potential of multi-modal foundation models and video diffusion through two key innovations: \textbf{1)} physics-conditioned flexible material behavior modeling, and \textbf{2)} optical flow-guided iterative refinement for high-fidelity 4D dynamic scene simulation (see Fig.~\ref{fig:pipeline}).  
A detailed comparison of the feasibility of our approach versus existing methods across various configurations is provided in Tab.~\ref{tab:meth}.

Our method leverages large pre-trained visual foundation models (\eg, GPT-4~\cite{achiam2023gpt}) to identify material types and initialize material parameters based on visual and text-based queries, ensuring that initial simulations are grounded in accurate material properties. 
Depending on the input type (\ie, multi-view images, dynamic scene videos, or a single image), we employ different scene reconstruction methods~\cite{kerbl3Dgaussians,wu2024-4dgs,smart2024splatt3r} to generate 3D Gaussian splats, providing a differentiable representation for simulation. The material type and initial material properties inferred from the large visual foundation model are then assigned to the 3D Gaussian splats of the simulated object, ensuring that the simulation starts with realistic physical characteristics. To refine these parameters automatically, we use optical flow as guidance within the video diffusion framework, integrated with a differentiable Material Point Method (MPM). 
Unlike render loss or SDS loss, which are limited to small and stable motions or are computationally intensive, optical flow loss offers a memory-efficient alternative that effectively captures large and complex motions.
This enables our method to optimize the material properties to reflect the interactions modeled in the generated video, allowing for simulations that are not only visually plausible but also physically accurate. Our approach supports a diverse range of material behaviors, thus enhancing the flexibility and precision of simulations to represent real-world scenarios comprehensively.
In summary, our contributions are as follows:
\begin{itemize}
    \item We present a novel multi-modal approach to infer 3D Gaussian splats along with the material type and initial material parameters to ensure accurate initial simulation.
    \item We introduce the guidance of optical flow to optimize material parameters through video diffusion models and differentiable MPM.
    \item Extensive experiments show that our approach achieves physically realistic 4D dynamic scene simulations on both synthetic and real-world scenes.
\end{itemize}

\section{Related Work}
\label{sec:liter}

\subsection{Dynamic 3D Reconstruction}
Dynamic 3D reconstruction has become a fundamental aspect of computer vision, enabling detailed representations of real-world scenes and objects for downstream tasks. Neural Radiance Fields (NeRF)~\cite{mildenhall2021nerf} pioneered high-fidelity 3D reconstruction~\cite{Ye2023IntrinsicNeRF, huang2024nerfdet++, tang2024nd, wang2024neurodin, ming2022idf, liu2021coxgraph, liu2023raydf, liu2025ddf} by learning continuous volumetric scene functions from posed images, achieving impressive visual quality. However, NeRF's high computational demands and reliance on multi-view input limit its effectiveness for dynamic scenes. To enhance real-time reconstruction and efficiency, methods leveraging Gaussian splats~\cite{kerbl3Dgaussians, tang2024hisplat, cui2024streetsurfgs, chen2024gigags} for representing surfaces in a differentiable and memory-efficient manner have been developed. Specifically, 3D Gaussian Splatting (3DGS)~\cite{kerbl3Dgaussians} and PGSR~\cite{chen2024pgsr} are designed for multi-view images with known poses, enabling high-quality static scene reconstruction. For scenarios involving 4D dynamic scene videos, 4D Gaussian Splatting (4DGS)~\cite{wu2024-4dgs, cai2024dynasurfgs} and Grid4D~\cite{xu2024grid4d} extend the Gaussian splatting concept to accommodate time-varying data. When only a single image is available, Splatt3R~\cite{smart2024splatt3r} adapts Gaussian splatting to reconstruct scenes without requiring camera parameters or depth information. These Gaussian splatting methods enable continuous representations adaptable to various input types, which is crucial for creating detailed and versatile scene representations, providing the foundation for accurate physics-based simulations in complex environments.

\subsection{Physics-Based Motion Generation}
Physics-based motion generation encompasses methods that integrate physical principles to simulate realistic object interactions. Differentiable simulation frameworks, such as the differentiable Material Point Method (MPM)~\cite{jiang2015mpm}, play a key role by allowing gradient-based optimization for motion generation tasks. These frameworks enable the fine-tuning of physical parameters and enhance simulations with realistic material behavior.

PAC-NeRF~\cite{li2023pac} combines NeRF with a differentiable MPM to simulate the physical properties of different materials, providing a degree of flexibility through its particle-based representation. However, its limited rendering fidelity restricts its effectiveness for high-complexity scenes. PhysGaussian~\cite{xie2024physgaussian} improves visual quality by coupling 3D Gaussian splatting with MPM, but it requires manual parameter setting, which limits its adaptability for diverse, real-world applications. Therefore, some existing methods~\cite{zhang2024physdreamer, huang2024dreamphysics, liu2024physics3d} utilize video diffusion priors to estimate the elasticity parameters of objects. While these approaches incorporate physics to some extent, they are typically constrained to learning only elasticity, limiting their ability to handle more diverse material behaviors like plasticity or viscosity. Furthermore, GIC~\cite{cai2024gic} leverages Gaussian representations for explicit shape modeling and continuum mechanics for implicit shape inference, but it requires multi-view input and known camera poses, making it less suited for real-world dynamic video data where geometry and material types are unknown.

These limitations highlight the need for automatic adjustment of material properties and handling of various material types. Our approach integrates optical flow-guided parameter refinement within a video diffusion framework, enhanced through differentiable MPM, to capture complex object interactions and adapt to diverse material types.

\subsection{Video Generation Models}
Video generation models have shown significant potential for simulating dynamic scenes due to their ability to create visually compelling sequences based on learned priors. These models~\cite{blattmann2023stable,yang2024cogvideox,niu2024mofa,meng2024phygenbench, Ye2024DiffPano,li2024seeground} take diverse prompts (\eg, text, image, trajectory, camera control, brush, and \textit{etc.}) and are trained on extensive datasets to implicitly capture relationships between object appearance and motion. 

To effectively utilize video generation results as guidance for physical-based simulation, Score Distillation Sampling (SDS)~\cite{poole2022dreamfusion} is commonly used to align generated content with target properties by distilling scores from large pre-trained models. However, SDS loss requires substantial computational resources and high memory usage. On the other hand, perceptual loss is used to maintain coherence and stability between video frames~\cite{zhang2024physdreamer}, ensuring that the generated motion remains smooth. This constraint forces the model to limit generated movements to small-scale changes, making it unsuitable for simulations involving large or rapid motions. Our approach addresses these limitations by leveraging video diffusion models to perform optical flow-guided refinement. 
% This enables the simulation of complex, physically accurate interactions that can accommodate larger and more varied motions, bridging the gap between visual realism and physical fidelity.

\section{Method}
\label{sec:method}
% Our approach leverages the power of multi-modal foundation models (\ie, Gaussian splatting models~\cite{chen2024pgsr,xu2024grid4d,smart2024splatt3r} and GPT-4~\cite{achiam2023gpt} and video diffusion model~\cite{yang2024cogvideox} to achieve high-fidelity 4D dynamic scene simulation. 
% The pipeline starts with 3D scene reconstruction to create a differentiable representation of the scene, followed by the initialization of material properties. These properties are then refine automatically using optical flow-guided optimization with a video diffusion framework, all integrated through a differentiable Material Point Method (MPM) for adaptive simulation. 

In this paper, we aim to automatically estimate material properties to enable high-fidelity 4D scene reconstruction from the input image(s) through physical simulation.
The overview of our pipeline is illustrated in Fig.~\ref{fig:pipeline}.
Given input image(s) $I$, which can be multi-view frames, video frames, or a single frame, our pipeline starts with inferring the 3D scene $\{\mathcal{G}_p\}$ using Gaussian splatting models. Meanwhile, we utilize GPT-4 with text prompts to identify the material type $M$, density $\rho$, and the initial material parameters ${M_p}_0$ for the object from the input image.
% (${M_p}_0 = \{E_0, \nu_0, {\tau_Y}_0, {\theta_{fric}}_0, {\eta}_0, \mu_0, \kappa_0\}$) for the object. 
With the 3D Gaussian splats $\{\mathcal{G}_p\}$ and the material properties, we simulate the object movements under the differentiable MPM with an applied input force. % $\mathbf{F}$. 
Over $t$ timesteps, we render the scene from a certain viewpoint $\mathbf{v}=[\mathbf{p}, \mathbf{d}]$ to generate several frames $\{\hat{I}_0,\ldots,\hat{I}_t\}$. We then employ the video diffusion model, along with the input image and related prompts, to produce a sequence of video frames $\{I_0,\ldots,I_t\}$ that capture the object motions. Distinguishing our approach from previous methods, we use the optical flows derived from the video frames, rather than the frames themselves, to guide the optimization of material parameters. We ultimately obtain the optimized material parameters $M_p$ and facilitate 4D reconstruction across various viewpoints.

\subsection{Preliminaries}

% \subsubsection{3D Scene Reconstruction} 
\noindent\textbf{3D Scene Reconstruction:} Following PhysGaussian~\cite{xie2024physgaussian}, our method reparameterizes the 3D scene using unstructured 3D Gaussian kernels, denoted by $\mathcal{G}_p = \{(\mathbf{x}_p, \sigma_p, \mathbf{A}_p, \mathbf{C}_p)\}_{p \in\mathcal{P}}$, where $\mathbf{x}_p$ represents the centers of the Gaussians, $\sigma_p$ the opacities, $\mathbf{A}_p$ the covariance matrices, and $\mathbf{C}_p$ the spherical harmonic coefficients. This explicit representation provides a differentiable and memory-efficient structure, enabling the modeling of complex scenes in various input settings.

To generate 3D Gaussian splats, we employ different methods based on the input type. For multi-view images with known camera poses, we utilize PGSR~\cite{chen2024pgsr} to construct a static 3D representation. For dynamic scene video $\{I_0,\ldots,I_t\}$, Grid4D~\cite{xu2024grid4d} is employed to produce a time-dependent 3D representation, capturing the temporal evolution of the scene. When dealing with a single image $I$, Splatt3R~\cite{smart2024splatt3r} infers the scene structure using learned priors to create 3D splats.
\vspace{\baselineskip}
% \subsubsection{Material Types and Properties}
% \subsubsection{Continuum Mechanics}

\noindent\textbf{Continuum Mechanics:} Continuum mechanics describes the motion and deformation of materials using a time-dependent continuous deformation map: $\mathbf{x} = \phi(\mathbf{X}, t)$, where $\mathbf{X}$ represents the undeformed material space and $\mathbf{x}$ denotes the deformed space at time $t$. The deformation gradient $\mathbf{F}(\mathbf{X}, t) = \nabla_{\mathbf{X}} \phi(\mathbf{X}, t)$ encodes the local transformations in material, including stretching, rotation, and shear.

Our work simulates seven types of materials: \textit{elastic, plasticine, metal, foam, sand, Newtonian fluids, and non-Newtonian fluids}. Each material type exhibits unique behaviors, necessitating different physical properties for accurate modeling. To ensure these properties are represented accurately, the evolution of $\phi$ must adhere to fundamental physical laws, including the conservation of mass and momentum. Conservation of mass ensures consistent density within material regions over time, while conservation of momentum is expressed through:
\begin{equation}
    \rho(\mathbf{x}, t) \dot{\mathbf{v}}(\mathbf{x}, t) = \nabla \cdot \boldsymbol{\sigma}(\mathbf{x}, t) + \mathbf{f}^{\text{ext}},
\end{equation}
where $\boldsymbol{\sigma}$ is the Cauchy stress tensor and $\mathbf{f}^{\text{ext}}$ denotes external forces. The deformation gradient $\mathbf{F}$ is decomposed into elastic $\mathbf{F}^\text{E}$ and plastic $\mathbf{F}^\text{P}$ components to account for permanent deformations. This decomposition is essential for simulating complex material behaviors, from the elasticity of foam to the plastic flow of metals, and enables our framework to model diverse real-world materials effectively.
\vspace{\baselineskip}

% \subsubsection{Material Point Method}
\noindent\textbf{Material Point Method:} Material Point Method (MPM) bridges the strengths of Lagrangian and Eulerian approaches by discretizing the material into particles, $\mathbf{P} = \{(\mathbf{x}_p, \mathbf{v}_p, \mathbf{F}_p)\}$, each representing a small region of the material. These particles track properties such as position $\mathbf{x}_p$, velocity $\mathbf{v}_p$, and deformation gradient $\mathbf{F}_p$. Lagrangian particles ensure mass conservation, while the Eulerian grid representation facilitates momentum conservation. The interaction between particles and the grid is handled using B-spline kernel functions. Momentum conservation is discretized over time steps, updating grid velocities and transferring them back to particles to adjust their positions:
\begin{equation}
    \mathbf{x}_p^{n+1} = \mathbf{x}_p^n + \Delta t \mathbf{v}_p^{n+1}.
\end{equation}
The deformation gradient $\mathbf{F}_p^{\text{E}}$ is updated to track changes, with adjustments made for plasticity as needed. This combination enables the simulation of complex deformations and interactions, essential for realistic material behavior in dynamic scenes.

\subsection{Material Property Initialization}
To simulate an object accurately, knowing the material type is essential for applying deformations that adhere to physical laws. Additionally, the initial values of material parameters play a significant role in the subsequent optimization process. Therefore, utilizing foundation models to estimate initial material properties is critical for achieving realistic simulations. In our work, we infer initial material properties by querying GPT-4~\cite{achiam2023gpt} with the input image and questions. As shown in Fig.~\ref{fig:pipeline}, the query involves identifying the closest material type $M$ and inferring its density $\rho$ along with the corresponding initial material parameters ${M_p}_0$. 
% Specifically, the material parameters for each material type are as follows:
% \begin{itemize}
%     \item \textit{Elastic:} Young's modulus $E$ and Poisson's ratio $\nu$.
%     \item \textit{Plasticine:} Young's modulus $E$, Poisson's ratio $\nu$, and yield stress $\tau_Y$.
%     \item \textit{Metal:}  Young's modulus $E$, Poisson's ratio $\nu$, and yield stress $\tau_Y$.
%     \item \textit{Foam:} Young's modulus $E$, Poisson's ratio $\nu$, and plastic viscosity $\eta$.
%     \item \textit{Sand:} friction angle $\theta_{fric}$.
%     \item \textit{Newtonian fluid:} fluid viscosity $\mu$ and bulk modulus $\kappa$.
%     \item \textit{Non-Newtonian fluid:} shear modulus $\mu$, bulk modulus $\kappa$, yield stress $\tau_Y$, and plastic viscosity $\eta$.
% \end{itemize}
More details of the material parameters for seven different material types can be found in Appendix~\ref{sec:supp_material}.

\subsection{Material Parameter Optimization with Optical Flow Guidance}
After the initial material properties are assigned (\eg, ${M_p}_0=\{E_0, \nu_0, {\tau_Y}_0, \eta_0\}$ in Fig.~\ref{fig:pipeline}), optimizing these parameters during the simulation is crucial for capturing realistic object interactions and ensuring adherence to physical laws. Our approach leverages optical flow to guide the optimization process dynamically, enhancing the material parameters as the simulation evolves.

Given an input image $I$ and a set of prompts $\mathcal{Q}$, our method generates a video input $V=\{I_0, \ldots, I_t\}$ using different video diffusion models. This flexibility allows the selection of prompts and models best suited to the simulation's needs. For instance, when only an image is provided, the Stable Video Diffusion (SVD) model~\cite{blattmann2023stable} is used. When both image and text prompts are available, CogVideoX~\cite{yang2024cogvideox} generates the video. For scenarios involving trajectory or camera motion, MotionCtrl~\cite{wang2024motionctrl} is employed, while for more complex motions involving brush strokes, MOFA-Video~\cite{niu2024mofa} or the Kling 1.5 Model~\cite{kling} are utilized.

Given the video input $V$, the optical flow $\mathbf{U}(I_t, I_{t+1})$ between consecutive frames is computed to track motion and detect discrepancies between the simulated and observed movements. This information is used to adjust the material parameters $M_p = \{E, \nu, \tau_Y, \eta\}$, ensuring that the simulation reflects realistic behavior over time.

The optimization process begins by analyzing the predicted motion from the simulation and comparing it to the optical flow-derived motion in the input frames. A flow-based loss $ \mathcal{L}_{\text{flow}}$ is minimized to align the simulation output $\hat{V}$ with the actual video $V$:
\begin{equation}
    \mathcal{L}_{flow} = \sum_{t} \|\mathbf{U}(I_t, I_{t+1}) - \hat{\mathbf{U}}(\hat{I}_t, \hat{I}_{t+1})\|^2,
\end{equation}
where $\hat{\mathbf{U}}$ denotes the flow derived from simulated frames. 

This process iteratively optimizes $M_p$ with the Moving Least Squares Material Point Method (MLS-MPM)~\cite{hu2018mlsmpm}. MLS-MPM extends traditional MPM by incorporating a smoother interpolation scheme that ensures better accuracy and stability, particularly during large deformations and complex material behaviors. 

Besides, unlike image-level render loss~\cite{zhang2024physdreamer}, which is restricted to subtle motions due to the instability and high noise in video generation outputs, or Score Distillation Sampling (SDS) loss~\cite{poole2022dreamfusion}, which inherently requires significant memory and becomes even more demanding when applied to video diffusion models with complex prompts, using optical flow loss as guidance presents a more efficient solution. It effectively captures motion while optimizing material parameters and conserves memory, making it a practical approach for high-fidelity simulations.

After optimizing the material parameters $M_p$, the updated properties are integrated into the simulation to control the motion and deformation of the 3D Gaussian splats $\{\mathcal{G}_p\}$ with the differentiable MPM framework. The final step involves rendering the simulation from new viewpoints $\{\mathbf{v}_1, \mathbf{v}_2, \ldots, \mathbf{v}_k\}$ for 4D reconstruction.

\section{Experiments}
\label{sec:exp}

In this section, we showcase the ability of our method to automatically optimize material parameters across a diverse set of material types, and evaluate its effectiveness on both synthetic and real-world datasets.

\subsection{Evaluation on Synthetic Dataset}

\begin{figure}[t]
  \centering
   \includegraphics[width=1.\linewidth]{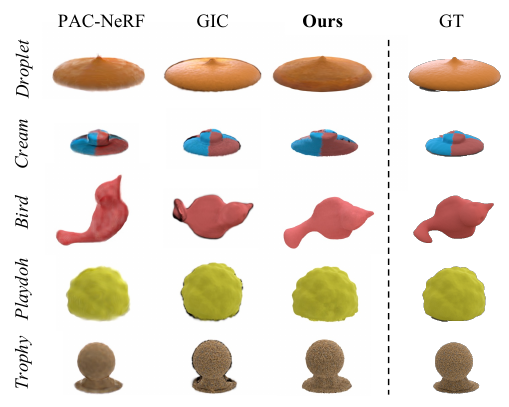}
   \vspace{-0.8cm}
   \caption{Qualitative results of all methods on synthetic dataset.}
   \label{fig:comp_syn}
   \vspace{-0.5cm}
\end{figure}

\noindent\textbf{Dataset:} We use the synthetic dataset introduced by PAC-NeRF~\cite{li2023pac}, which features 9 objects made of different materials, including elastic (\textit{Torus}, \textit{Bird}), plasticine (\textit{Playdoh}, \textit{Cat}), sand (\textit{Trophy}), and both Newtonian (\textit{Droplet}, \textit{Letter}) and non-Newtonian fluids (\textit{Cream}, \textit{Toothpaste}). Each object includes RGB images from 11 distinct viewpoints, with 10 to 16 frames per viewpoint. Since our work focuses on system identification and validation of our effectiveness in optimizing material parameters, we adopt the 3D Gaussian splats generated from the first frame of dynamic reconstruction by GIC~\cite{cai2024gic} as our simulation objects. Besides, we choose the RGB images of a certain viewpoint from the dataset for each object instead of the video diffusion output in our pipeline for a fair comparison.

\noindent\textbf{Baselines:} For comparing system identification performance, we evaluate our method employing PAC-NeRF~\cite{li2023pac} and GIC~\cite{cai2024gic} as baselines. PAC-NeRF is designed for system identification tasks and combines neural representations with differentiable physics for material parameter estimation. GIC focuses on dynamic 3D reconstructions using Gaussian splatting to capture shape and motion. Both baselines represent state-of-the-art approaches in this domain, offering a robust comparison for assessing our method’s effectiveness. For a fair comparison, all baselines and our method optimize parameters starting from the same initial settings introduced in PAC-NeRF.

\noindent\textbf{Metric:} We evaluate \textbf{1)} Relative Error ($\mathbf{RE}=|M_p-\hat{M_p}|/M_p$) comparing predicted versus ground-truth material parameters and \textbf{2)} End-Point Error (\textbf{EPE})~\cite{flownet} quantifying optical flow deviations to measure the system identification performance across various material types.

% \vspace{\baselineskip}
\noindent\textbf{Analysis:} We report the RE and EPE for system identification performance across various material parameters on the synthetic dataset (Tab.~\ref{tab:comp_syn}) for PAC-NeRF, GIC, and our \paperName{}. Our method consistently achieves lower RE and EPE than PAC-NeRF in most cases and remains competitive with GIC, demonstrating its effectiveness in predicting material properties. Specifically, for \textit{Droplet} and \textit{Letter}, our approach yields the lowest RE and EPE, indicating strong performance in \textit{Newtonian fluids}. For \textit{non-Newtonian fluids} like \textit{Toothpaste}, it also delivers competitive results, maintaining low shear and bulk modulus errors.  
For \textit{Trophy}, which represents \textit{sand} (\ie, granular material), our method achieves the lowest error in friction angle estimation, highlighting its precision in granular material properties. While system identification remains challenging for complex materials, our approach performs comparably to baseline methods. As shown in Fig.~\ref{fig:comp_syn}, our method also produces deformations that closely match the ground truth and preserves sharper textures across different material types.

% --------------------------------------------------

\subsection{Evaluation on Real-world Dataset}

\begin{table}[tp]\tabcolsep=0.2cm
    \centering
    \resizebox{1.\linewidth}{!}{
    \begin{tabular}{lcc}
    \toprule
    \textbf{Method} & \textbf{Physical-realism} $\uparrow$ & \textbf{Photo-realism} $\uparrow$ \\ \hline
    PhysGaussian~\cite{xie2024physgaussian} & 2.67 & 2.91 \\
    PhysDreamer~\cite{zhang2024physdreamer} & 2.58 & 2.91 \\
    Physics3D~\cite{liu2024physics3d} & 2.64 & 2.98 \\ 
    % PhysGen~\cite{liu2025physgen} & - & - \\
    \textbf{Ours} & \textbf{3.44} & \textbf{3.53} \\
    \bottomrule
    \end{tabular}
    }
    \vspace{-0.3cm}
    \caption{Human evaluation on real-world dataset.}
    \vspace{-0.3cm}
\label{tab:comp_real}
\end{table}

\begin{table}[tp]\tabcolsep=0.3cm
\centering
\resizebox{1.\linewidth}{!}{
\begin{tabular}{cccc}
\toprule
\textbf{PhysGaussian} & \textbf{PhysDreamer} & \textbf{Physics3D} & \textbf{Ours}\\ \hline
3.70 & 3.87 & 4.41 & \textbf{3.08}  \\
\bottomrule
\end{tabular}
}
\vspace{-0.3cm}
\caption{Evaluation metric (ECMS$\downarrow$) on real-world dataset.}
\label{tab:comp_real_ecms}
\vspace{-0.5cm}
\end{table}

\noindent\textbf{Dataset:} Our real-world evaluation includes a comprehensive set of scenes to cover diverse material types. We use four publicly available scenes from PhysDreamer~\cite{zhang2024physdreamer} (\textit{Alocasia}, \textit{Carnation}, \textit{Hat}, and \textit{Telephone}), along with additional scenes from other sources: \textit{Fox} from InstantNGP~\cite{muller2022instant}, \textit{Plane} from NeRFStudio~\cite{nerfstudio}, and \textit{Kitche}n from Mip-NeRF 360~\cite{barron2022mip}. To ensure coverage of all seven material types, we introduce two additional self-collected scenes, \textit{Jam} and \textit{Sandcastle}.

\noindent\textbf{Baselines:} For real-world simulation, we compare our method with the following baselines: PhysGaussian~\cite{xie2024physgaussian}, PhysDreamer~\cite{zhang2024physdreamer}, and Physics3D~\cite{liu2024physics3d}.
PhysGaussian combines 3D Gaussian splatting with differentiable MPM for high-fidelity rendering and accurate physical simulations but requires manual parameter settings. PhysDreamer and Physcis3D leverage priors from video diffusion models to estimate material properties, but are limited to elastic materials. To evaluate our method, we apply larger forces than previous baselines. For material settings, we use GPT-queried materials for PhysGaussian, maintain elasticity for PhysDreamer, and incorporate both elasticity and viscoelasticity for Physics3D. PhysGen~\cite{liu2025physgen} is not included in this comparison due to its reliance on perceptual information (\eg, albedo), which is unavailable in real-world scenarios. Instead, we provide a separate evaluation on its synthetic dataset, with additional implementation details in Appendix~\ref{sec:supp_exp_single}.

\noindent\textbf{Metric:} To evaluate performance, we conduct a human evaluation where 32 participants rate 36 videos generated by different methods, presented in random order. Each video is rated on a 5-point Likert scale, from 1 (strongly disagree) to 5 (strongly agree), based on two criteria: \textbf{physical-realism} and \textbf{photo-realism}. Participants independently assess each video along these two dimensions to provide a comprehensive evaluation of simulation realism. Furthermore, to assess motion quality, we consider the \textit{Energy Minimization Principle} for physically plausible motion and compute the Energy-Constrained Motion Score (\textbf{ECMS}):
\begin{equation}
 E = \sum_{t} \|\mathbf{v}_{t+1} - \mathbf{v}_t\|^2 + \|\nabla^2 \mathbf{v}_t\|^2 + \alpha \frac{1}{\sum_{t} \|\mathbf{v}_t\|}
\end{equation}
where $\mathbf{v}_t$ denotes velocity and $\alpha=0.1$.

% \vspace{\baselineskip}
\noindent\textbf{Analysis:} To thoroughly evaluate robustness, we applied large input forces in several scenes to induce substantial object deformations and motions, challenging the capacity to accurately simulate intense physical dynamics. As shown in Tab.~\ref{tab:comp_real}, our method achieves the highest average scores for both physical-realism and photo-realism, significantly outperforming the baseline methods on the real-world dataset. Specifically, our method is the only method achieving scores above 3, while all baseline methods fall below this threshold. And our method achieves the lowest ECMS as shown in Tab.~\ref{tab:comp_real_ecms}, indicating more plausible motion. The results underscore the limitations of existing methods in capturing complex physical interactions under challenging scenarios.

\begin{figure*}[ht]
  \centering
   \includegraphics[width=.95\linewidth]{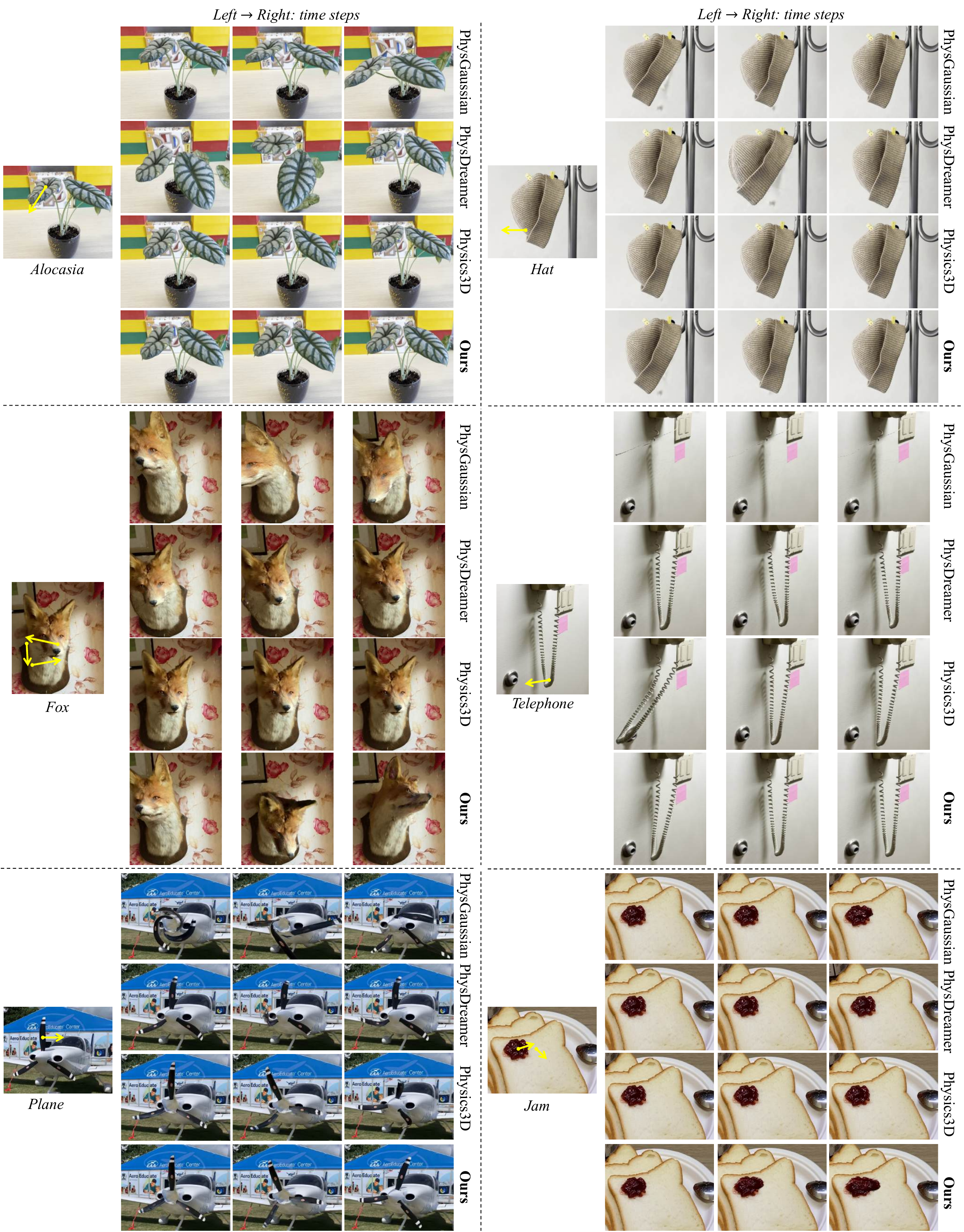}
   \caption{Qualitative results of all methods on real-world dataset. The yellow arrows show the input force for the simulated objects.}
   \label{fig:comp_real}
\end{figure*}

The qualitative results in Fig.~\ref{fig:comp_real} further illustrate the advantages of our method. Even under high-force conditions, our approach produces realistic and stable simulations, capturing both material deformation and dynamic interactions with greater fidelity compared to the baselines. These results highlight the robustness and adaptability of our method for scenes involving diverse materials and large motions, demonstrating the effectiveness of integrating multi-modal foundation models with optical flow-guided optimization.

Additional implementation details are provided in Appendix~\ref{sec:supp_implement}, and more visual comparisons are available in Appendix~\ref{sec:supp_syn_real}.

% --------------------------------------------------

\begin{table*}[htp]\tabcolsep=0.8cm
\centering
\resizebox{1.\linewidth}{!}{
\begin{tabular}{lccccc}
\toprule
\textbf{Object} & \textbf{PAC-NeRF~\cite{li2023pac}} & \textbf{GIC~\cite{cai2024gic}} & \textbf{Ours} ($\mathcal{L}_{flow}$) & $\mathcal{L}_{render}$ & $\mathcal{L}_{sds}$ \\ \hline
Droplet & \underline{0.06} / \underline{0.28} & 0.41 / 0.42 & \textbf{0.02} / \textbf{0.08} & 0.48 / 0.63 & 0.98 / 1.01 \\ 
Letter  & \underline{0.26} / \underline{0.11} & \textbf{0.02} / \textbf{0.07} & 0.46 / 0.13 & 0.58 / 0.20 & 0.70 / 0.18 \\ 
\hdashline
Cream      & 3.04 / 0.61 & \textbf{0.21} / \textbf{0.20} & 0.54 / 0.64 & \underline{0.45} / \underline{0.30} & 1.27 / 2.76 \\
Toothpaste & 0.42 / 0.15 & \textbf{0.11} / \textbf{0.07} & \underline{0.24} / \underline{0.16} & 0.30 / 0.20 & \underline{0.24} / 0.19\\ 
\hdashline
Torus & \underline{0.06} / \underline{0.52} & \textbf{0.01} / \textbf{0.24} & 0.51 / 1.04 & 0.41 / 1.64 & 0.57 / 1.58 \\ 
Bird  & \underline{0.08} / \underline{0.62} & \textbf{0.04} / \textbf{0.36} & 0.31 / 0.69 & 0.37 / 0.88 & 0.76 / 3.07 \\ 
\hdashline
Playdoh & 0.37 / 0.55 & \textbf{0.10} / \textbf{0.16} & \underline{0.15} / \underline{0.29} & 0.54 / 4.02 & 0.65 / 4.05 \\ 
Cat     & \underline{0.31} / 0.34 & \textbf{0.02} / \textbf{0.08} & 0.34 / \underline{0.24} & 0.37 / 0.94 & 0.85 / 0.70 \\ 
\hdashline
Trophy & 0.10 / 3.30 & \underline{0.05} / \underline{1.85} & \textbf{0.01} / \textbf{1.33} & 0.17 / 3.15 & 0.30 / 3.58 \\ 
\hline
\textbf{Avg.} & 0.52 / 0.72 & \textbf{0.11} / \textbf{0.38} & \underline{0.29} / \underline{0.51} & 0.41 / 1.33 & 0.70 / 1.90 \\
\bottomrule
\end{tabular}
}
\vspace{-0.3cm}
\caption{System identification performance (RE$\downarrow$ / EPE$\downarrow$) on the synthetic dataset.}
\label{tab:comp_syn}
\vspace{-0.6cm}
\end{table*}

% ---------------------------

\begin{table}[t]\tabcolsep=0.3cm
\centering
\resizebox{1.\linewidth}{!}{
\begin{tabular}{cccc}
\toprule
\textbf{w/o GPT Init.} & \textbf{GPT Init. w/o Optim.} & \textbf{$\mathcal{L}_{render}$} & \textbf{Ours}\\ \hline
3.91 & 3.23 & 3.21 & \textbf{3.08} \\
\bottomrule
\end{tabular}
}
\vspace{-0.3cm}
\caption{Ablation results (ECMS$\downarrow$) on real-world dataset.}
\label{tab:abla_real_ecms}
\vspace{-0.5cm}
\end{table}

\subsection{Ablation Study}\label{sec:abla}

\begin{figure}[t]
\vspace{-0.2cm}
  \centering
   \includegraphics[width=\linewidth]{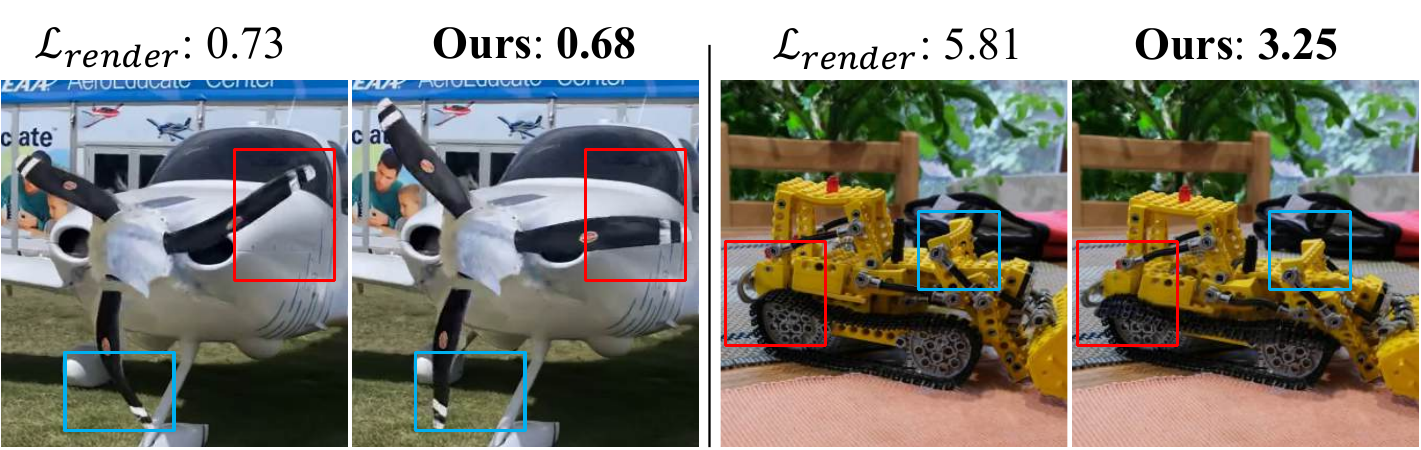}
   \vspace{-0.6cm}
   \caption{Comparisons of $\mathcal{L}_{render}$ and ours ($\mathcal{L}_{flow}$) with ECMS$\downarrow$.}
   \label{fig:abla_loss_render}
    \vspace{-0.5cm}
\end{figure}

We conduct a comprehensive ablation study to evaluate the contribution of key components in our method, using both synthetic and real-world datasets to provide a thorough understanding of their impacts on performance. 

\noindent\textbf{Effectiveness of Optical Flow Guidance:}  
We evaluate the impact of different loss functions on the synthetic dataset to highlight the significance of our optical flow guidance. Specifically, we compare the render loss $\mathcal{L}_{render} = \sum_t \lambda L_1(I_t,\hat{I}_t) + (1-\lambda)L_{D-SSIM}(I_t,\hat{I}_t)$~\cite{zhang2024physdreamer}, the SDS loss (with text prompt) $\mathcal{L}_{sds}$~\cite{poole2022dreamfusion}, and our proposed flow loss $\mathcal{L}_{flow}$.  
As shown in Tab.~\ref{tab:comp_syn}, optical flow guidance achieves lower RE and EPE across most material parameters, particularly viscosity $\eta$, bulk modulus $\kappa$, and yield stress $\tau_Y$ (detailed comparisons for each parameter are provided in Appendix~\ref{sec:supp_abla}), leading to more accurate material parameter optimization. These results demonstrate that optical flow guidance not only improves parameter estimation but also ensures more realistic motion in system identification compared to other guidance losses. 
Besides, Fig.~\ref{fig:abla_loss_render} and Tab.~\ref{tab:abla_real_ecms} show that $\mathcal{L}_{flow}$ performs better under large force applications and motions on real-world scenes.

\begin{figure}[t]
  \centering
   \includegraphics[width=1.\linewidth]{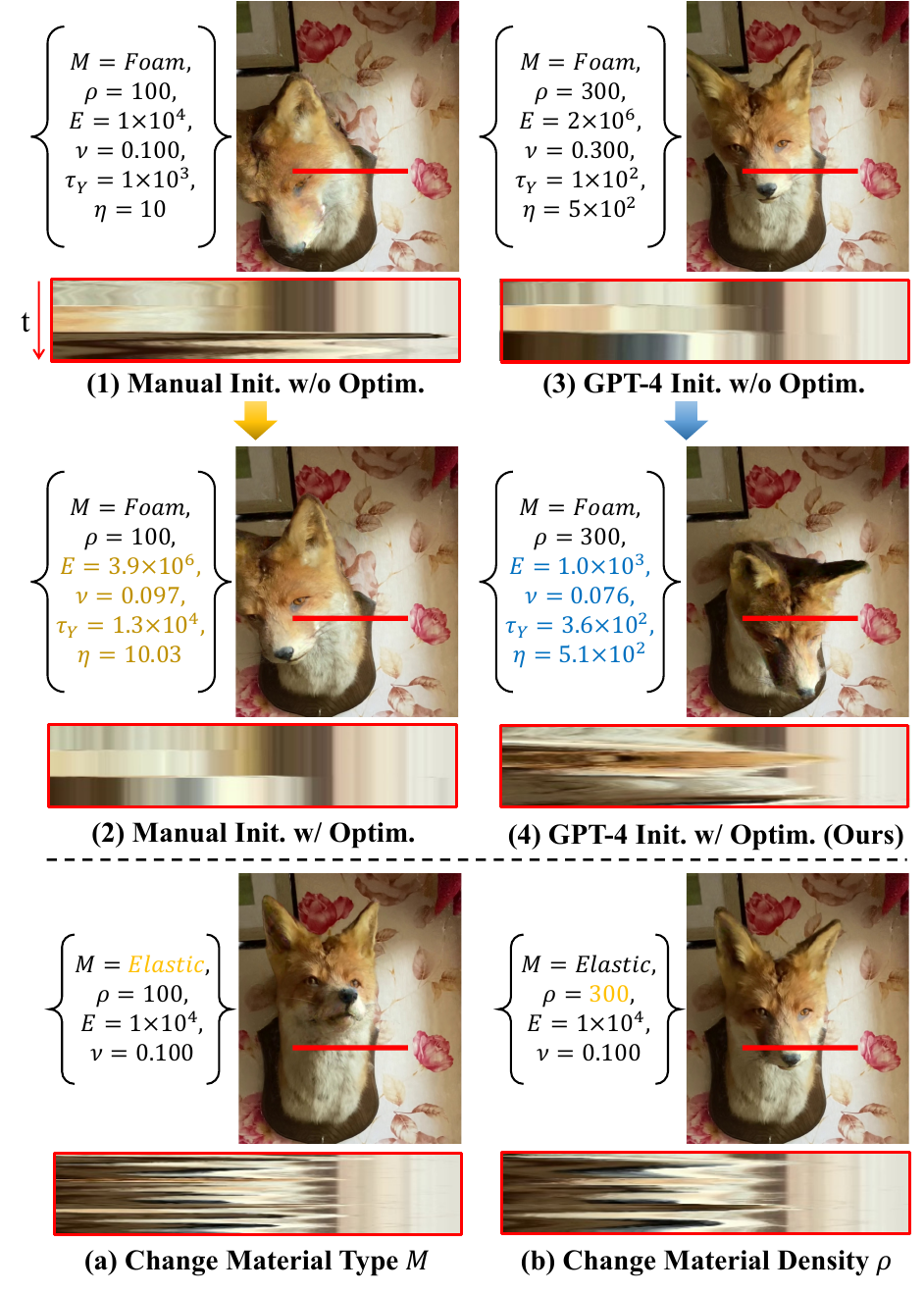}
   \vspace{-0.8cm}
   \caption{Ablations on physics reasoning, showing material values, timestep 30 frame, and deformation frequency.}
   \label{fig:abla_param}
    \vspace{-0.5cm}
\end{figure}

\noindent\textbf{Effect of Physics Reasoning with Foundation Model:} \label{sec:abla_gpt}
To evaluate the impact of using a foundation model (\eg, GPT-4) for physics reasoning, we conduct simulations on the real-world dataset with four configurations: \textbf{(1)} manually defined initial material properties without optimization, \textbf{(2)} use manually defined material properties as initialization followed by optimization, \textbf{(3)} use foundation model-inferred properties without optimization, \textbf{(4)} initialize material properties with foundation model-inferred values followed by optimization.
As shown in Fig.~\ref{fig:abla_param} and Tab.~\ref{tab:abla_real_ecms}, simulations initialized with manually defined values produce motions that partially reflect the effects of the input force but lack sufficient fidelity. Further optimization of these manually set values results in overly rigid or tense parameters, leading to less realistic outcomes. In contrast, simulations using foundation model-inferred values without additional optimization provide a more balanced starting point. By initializing with foundation model predictions and optimizing, our method achieves significantly more accurate material dynamics and visually convincing simulations.

In addition to parameter initialization, using a foundation model to identify material type and density ensures more realistic simulations. Accurate material classification directly affects an object’s physical response, as shown in Fig.~\ref{fig:abla_param} (a) and (b), where incorrect material properties lead to unrealistic outcomes. Leveraging the foundation model predictions enhances simulation fidelity, allowing for more accurate and consistent object behavior in complex scenes.

\section{Conclusion}
\label{sec:sum}

In this paper, we present a novel approach leveraging multi-modal foundation models and video diffusion with optical flow guidance for 4D dynamic scene simulation. Our method optimizes material parameters using optical flow guidance and integrates inferred properties for realistic simulations across diverse material types. Through comprehensive experiments on synthetic and real-world datasets, we demonstrate that our method outperforms existing baselines in accuracy and adaptability, showing its potential for enhancing physical realism in dynamic simulations.

% \noindent\textbf{Limitation:}  Our method deforms 3D Gaussian splats in simulation and renders results without relighting, which could be explored in future work to enhance visual realism.

\noindent{\textbf{Acknowledgement:}}
The work described in this paper was supported by grants from the Research Grants Council of the Hong Kong Special Administrative Region, China (Project No. GRF/PolyU 15606321 and Project No. GRF/PolyU 15607922). 

{
    \small
    \bibliographystyle{ieeenat_fullname}
    \bibliography{main}
}

% WARNING: do not forget to delete the supplementary pages from your submission 
\clearpage
\setcounter{page}{1}
\maketitlesupplementary
\renewcommand{\thesection}{\Alph{section}}
\setcounter{section}{0}

\section{Material Properties}\label{sec:supp_material}
In our work, we apply the Material Point Method (MPM) to simulate seven distinct material types: \textit{elastic, plasticine, metal, foam, sand, Newtonian fluid, and non-Newtonian fluid}. The definitions for the first five types are derived from PhysGaussian, while the latter two types follow the specifications outlined in PAC-NeRF. The material parameters for each type are detailed as follows: 
\begin{itemize}
    \item \textit{Elastic:} Young's modulus $E$ and Poisson's ratio $\nu$.
    \item \textit{Plasticine:} Young's modulus $E$, Poisson's ratio $\nu$, and yield stress $\tau_Y$.
    \item \textit{Metal:}  Young's modulus $E$, Poisson's ratio $\nu$, and yield stress $\tau_Y$.
    \item \textit{Foam:} Young's modulus $E$, Poisson's ratio $\nu$, and plastic viscosity $\eta$.
    \item \textit{Sand:} friction angle $\theta_{fric}$.
    \item \textit{Newtonian fluid:} fluid viscosity $\mu$ and bulk modulus $\kappa$.
    \item \textit{Non-Newtonian fluid:} shear modulus $\mu$, bulk modulus $\kappa$, yield stress $\tau_Y$, and plastic viscosity $\eta$.
\end{itemize}

For the synthetic dataset, each object's material type is predefined, while the real-world dataset lacks this specification. To infer the material types, we leverage a large pre-trained visual foundation model (\eg, GPT-4) for classification. As shown in Tab.~\ref{tab:supp_real}, some inferred material types differ from those specified in the original dataset (\ie, \textit{Alocasia} and \textit{Carnation} are assigned as \textit{elastic} in PhysDreamer). This underscores the importance of leveraging a robust visual foundation model to accurately determine material types, ensuring that simulations are based on reliable initial properties. Besides, we adopt the constitutive models for different material types as detailed in~\cite{zong2023neural, xie2024physgaussian, li2023pac}, which provide a strong framework for simulating a variety of materials with distinct physical behaviors.

% -------------------------
\section{Implementation Details}\label{sec:supp_implement}

% \subsection{Constitutive Models}
% % Table of constitutive models for each material type
% % detailed model functions can refer to the cite in PhysGaussian and PAC-Nerf

\subsection{Optimization Details}
In our experiments, we implement the differentiable MPM using Warp~\cite{warp}. Besides, we use RAFT~\cite{teed2020raft} to compute optical flow, which serves as a critical component for guiding the optimization process. For the real-world dataset, we segment the foreground points where input forces are applied using SAM2Point~\cite{guo2024sam2point}, ensuring precise identification of the regions impacted by force. Specific forces and durations are detailed in Tab.~\ref{tab:realworld_force}. Video generation is facilitated by CogVideoX~\cite{yang2024cogvideox}, which leverages a text prompt (refer to Tab.~\ref{tab:supp_real}) combined with a selected frame from the dataset to guide the synthesis process. All optimization tasks are performed on a single NVIDIA A800 GPU.

\subsection{Baselines} % implementation details
\noindent\textbf{PAC-NeRF/GIC:} PAC-NeRF and GIC estimate material parameters from multi-view images with deforming objects, while our approach relies solely on the 3D Gaussian splats captured at the initial frame. To ensure fair comparisons, we align the simulation timestep with those methods and use the deformed Gaussian splats at each timestep as additional supervision, employing Chamfer distance~\cite{fan2017chamfer} as the geometry loss.

\noindent\textbf{PhysGaussian/PhysDreamer/Physics3D:} PhysGaussian utilizes a fixed set of manually defined material properties, including material type, density, and associated parameters, without further optimization. This setup is comparable to the conditions outlined in Sec.~\ref{sec:abla} (1). Changes in material type, density, or further parameter optimization can lead to different simulation outcomes. To highlight the benefits of incorporating a large pre-trained visual foundation model for physics reasoning and parameter refinement, we apply the same input force to each scene and compare our results with PhysGaussian. While PhysDreamer and Physics3D include methods for parameter optimization, they are limited to the \textit{elastic} material type. For consistency and fair comparison, we assign \textit{elastic} as the material type for PhysDreamer and \textit{elastic with viscoelasticity} for Physics3D. Besides, in our experiments, we apply larger forces (refer to Tab.~\ref{tab:realworld_force}) than previous baselines on the real-world dataset. The weaker performance of baselines under these conditions highlights our approach's advantage in leveraging GPT initialization and optical flow guidance.

\noindent\textbf{PhysGen~\cite{liu2025physgen}:} Compared to other baselines, PhysGen utilizes foundation model-based physics reasoning, which removes the need for manual parameter initialization. However, its dependence on single-image input and lack of 4D reconstruction capability limit its applicability for comprehensive simulations.  Additionally, PhysGen simulates objects at a fixed depth, which restricts its ability to handle complex scenarios. These limitations prevent direct comparisons with other baselines on both the synthetic and real-world datasets. Therefore, we conduct a separate evaluation of our method on the dataset proposed by PhysGen to assess our simulation performance in Sec.~\ref{sec:supp_exp_single}.

\begin{table*}[ht]%\tabcolsep=0.1cm
    \centering
\resizebox{.75\linewidth}{!}{
    \begin{tabular}{lll}
        \toprule
        \textbf{Scene} & \textbf{Material Type} & \textbf{Text Prompt} \\
        \midrule
        Alocasia   & Foam    & \textit{The alocasia is swaying in the wind.}\\
        Carnation  & Foam    & \textit{The carnation is swaying in the wind.} \\
        Hat        & Elastic & \textit{The hat is given a tug.} \\
        Telephone  & Elastic & \textit{The telephone coil is given a tug.} \\
        Fox        & Foam    & \textit{The fox is shaking its head.} \\
        Plane      & Metal   & \textit{The propeller is spinning.} \\
        Kitchen    & Plasticine & \textit{The Lego on the table is being squeezed by a downward force.} \\
        Jam        & Non-Newtonian fluid & \textit{The jam on the toast is being spread.} \\
        Sandcastle & Sand    & \textit{The sandcastle on the beach is collapsing.} \\
    \bottomrule
    \end{tabular}
    }
    \vspace{-0.2cm}
    \caption{Material types inferred by GPT-4 and text prompts to generate guidance videos for scenes in real-world dataset.}
    \label{tab:supp_real}
\end{table*}

\begin{table*}[t]\tabcolsep=0.3cm
\centering
\resizebox{.6\linewidth}{!}{
\begin{tabular}{lll}
\toprule
\textbf{Scene} & \textbf{Force (s)} \textit{\ ($(x,y,z)$ for most cases)} & \textbf{Duration (s)} \\ \hline
Alocasia & $(0.25, 0, 0)$ & 1 \\
Carnation & $(-0.1,0,0)$  & 1 \\
Hat & $(1,-2,1)$ & 1\\
Telephone & $(-1,0,0)$ & 2\\
Fox & $(0,-0.5,0.25)\xrightarrow{}(0,0,-0.5)\xrightarrow{}(0,0.5,0.25)$ & 1$\xrightarrow{}$1$\xrightarrow{}$1 \\
Plane & $rotation\ scale=-10\xrightarrow{}-5$ & 0.8$\xrightarrow{}$1 \\
Kitchen & $(0,0,0.1)$ & 1 \\
Jam & $(0.2,0,0)\xrightarrow{}(0.1,0.2,0)$& 2$\xrightarrow{}$1 \\
Sandcastle & $release\ n_{layer}=50$& - \\
\bottomrule
\end{tabular}
}
\vspace{-0.3cm}
\caption{Simulation forces and durations on real-world dataset.}
\label{tab:realworld_force}
\vspace{-0.3cm}
\end{table*}

% -------------------------
\section{Additional Experimental Results}\label{sec:supp_experiment}

\begin{figure*}[ht]
    \centering
    \includegraphics[width=1.0\linewidth]{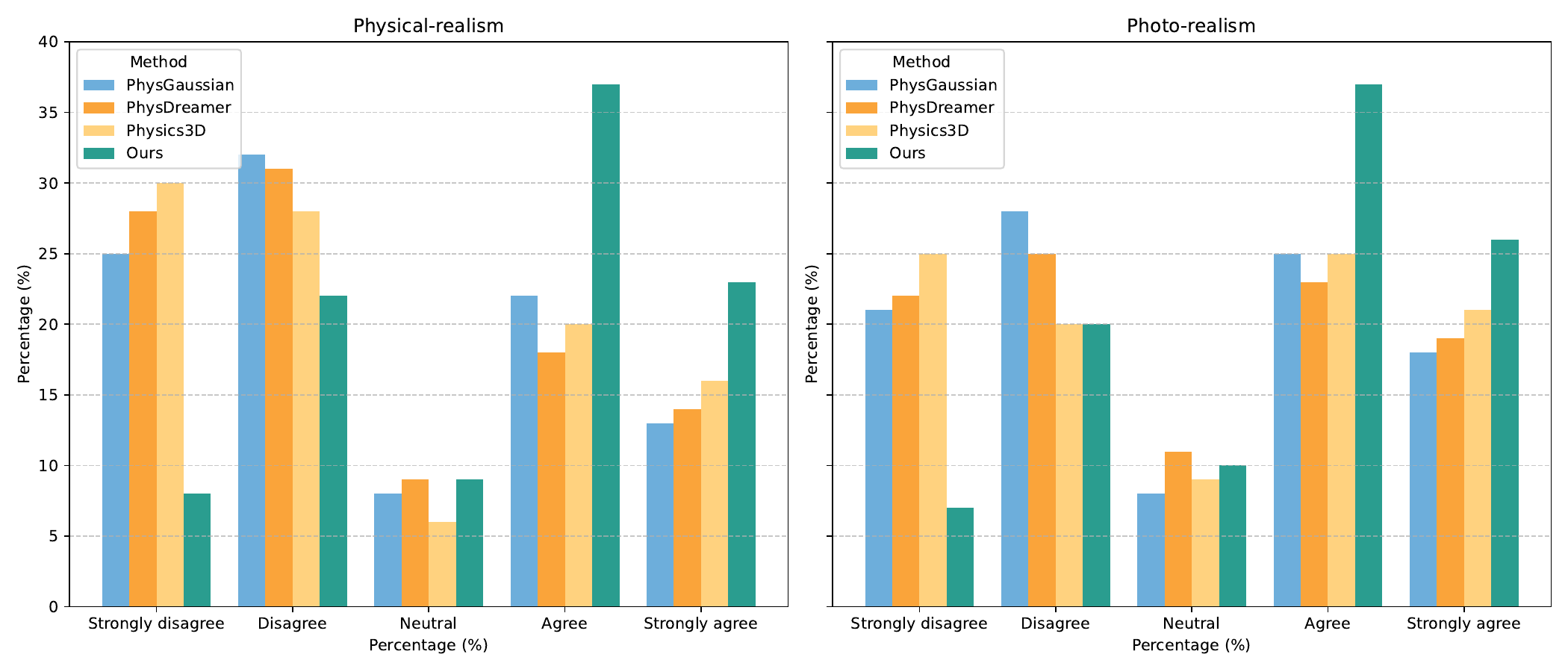}
    \vspace{-0.8cm}
    \caption{Human evaluation score distribution. The chart shows the score distribution for physical-realism and photo-realism based on human evaluations. Our method significantly outperforms the baseline methods across both metrics. The average score for our method approaches "Agree" for both criteria, indicating superior performance in producing realistic simulations.}
    \label{fig:supp_comp_eval}
\end{figure*}

\subsection{Human Evaluation}
We conducted a human evaluation to assess the physical-realism and photo-realism of videos generated by our method compared to the baseline methods: PhysGaussian, PhysDreamer, and Physics3D. As illustrated in Fig.~\ref{fig:supp_comp_eval}, our method achieves significantly higher scores in the "Agree" and "Strongly Agree" categories across both evaluation criteria. In contrast, baseline methods, such as PhysGaussian and PhysDreamer, received lower ratings, particularly in the "Neutral" and "Disagree" categories. The average score for our method surpasses 3.0, consistently approaching "Agree," reflecting its ability to simulate visually and physically plausible scenarios. This result underscores the robustness and effectiveness of our approach in addressing the challenges of material parameter optimization and dynamic scene simulation.

\subsection{Additional Experimental Results of Synthetic and Real-world Datasets}\label{sec:supp_syn_real}
\begin{figure}[t]
    \centering
    \includegraphics[width=1.0\linewidth]{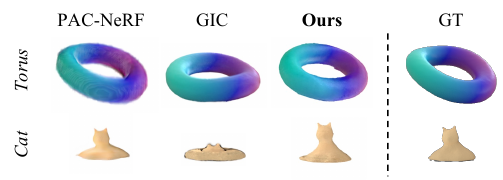}
    \vspace{-0.8cm}
    \caption{Qualitative results of all methods on synthetic dataset.}
    \label{fig:supp_comp_syn}
\end{figure}

\begin{figure}[t]
    \centering
    \includegraphics[width=1.0\linewidth]{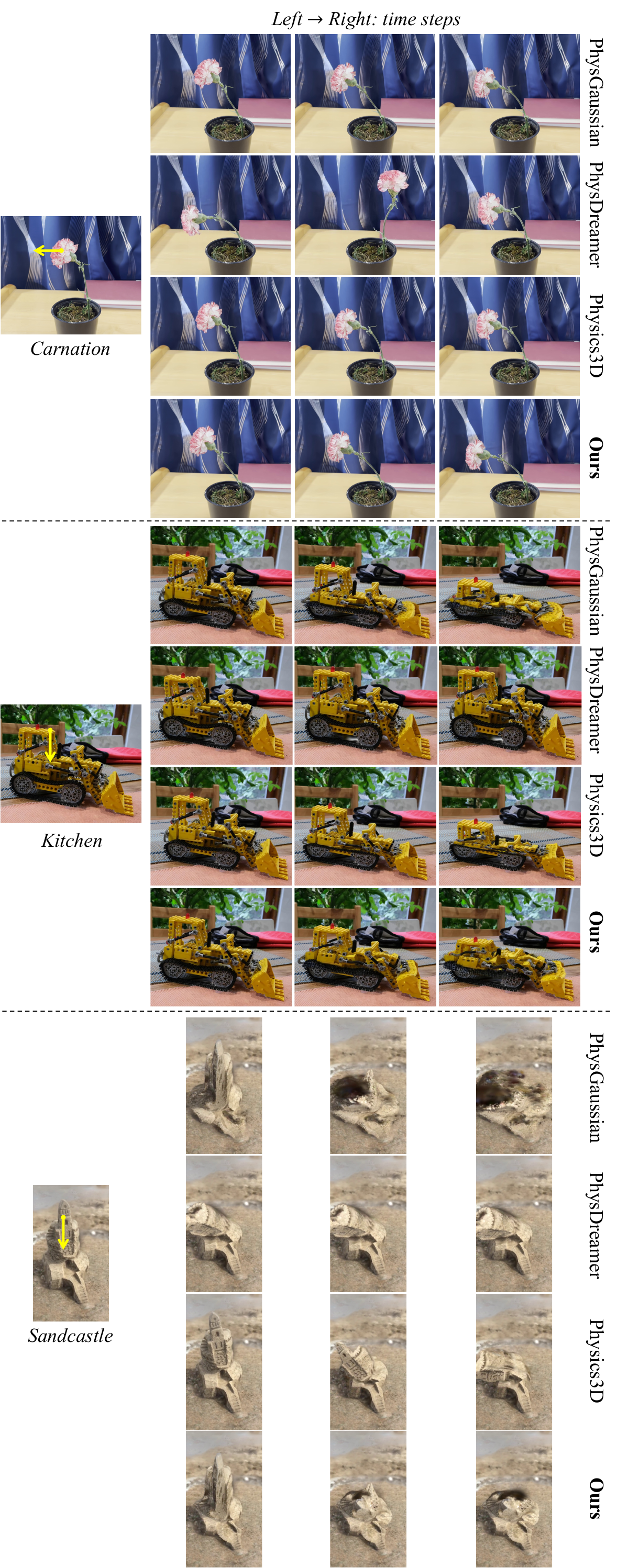}
    \vspace{-0.6cm}
    \caption{Qualitative results of all methods on real-world dataset. The yellow arrows show the input force for the simulated objects.}
    \label{fig:supp_comp_real}
\end{figure}

% link to a html of project page
% - comparisons with baselines on syn dataset
% - comparisons with baselines on real-world dataset
We provide additional comparisons between our method and baselines on both synthetic (see Tab.~\ref{tab:supp_comp_syn} and Fig.~\ref{fig:supp_comp_syn}) and real-world (see Fig.~\ref{fig:supp_comp_real}) datasets. These results emphasize the effectiveness of our method in handling diverse material behaviors and complex scenarios. To further enhance understanding and visualization, we offer more results in videos accessible through our project page: 
\href{https://zhuomanliu.github.io/PhysFlow}{https://zhuomanliu.github.io/PhysFlow}

% --------------------------

\begin{table}[htp]\tabcolsep=1cm
\centering
\resizebox{1.\linewidth}{!}{
\begin{tabular}{ccc}
\toprule
\textbf{PhysGen} & \textbf{PhysGaussian} & \textbf{Ours} \\ \hline
\textbf{0.54} & 2.95 & \underline{0.85}  \\
\bottomrule
\end{tabular}
}
\vspace{-0.3cm}
\caption{Evaluation metric (ECMS$\downarrow$) on PhysGen scenes.}
\label{tab:comp_physgen}
\vspace{-0.5cm}
\end{table}

\subsection{Additional Evaluation Using Single-view Input}\label{sec:supp_exp_single}
In addition to experiments on synthetic and real-world datasets, we further validate our method on scenarios with single-view input. Specifically, we employ Splatt3R to reconstruct 3D Gaussian splats from a single image and utilize an inpainting technique to recover the background geometry. For simulation, we focus exclusively on the segmented foreground points, ensuring the deformation and dynamics are isolated to the target object.

\begin{figure}[t]
    \centering
    \includegraphics[width=1.0\linewidth]{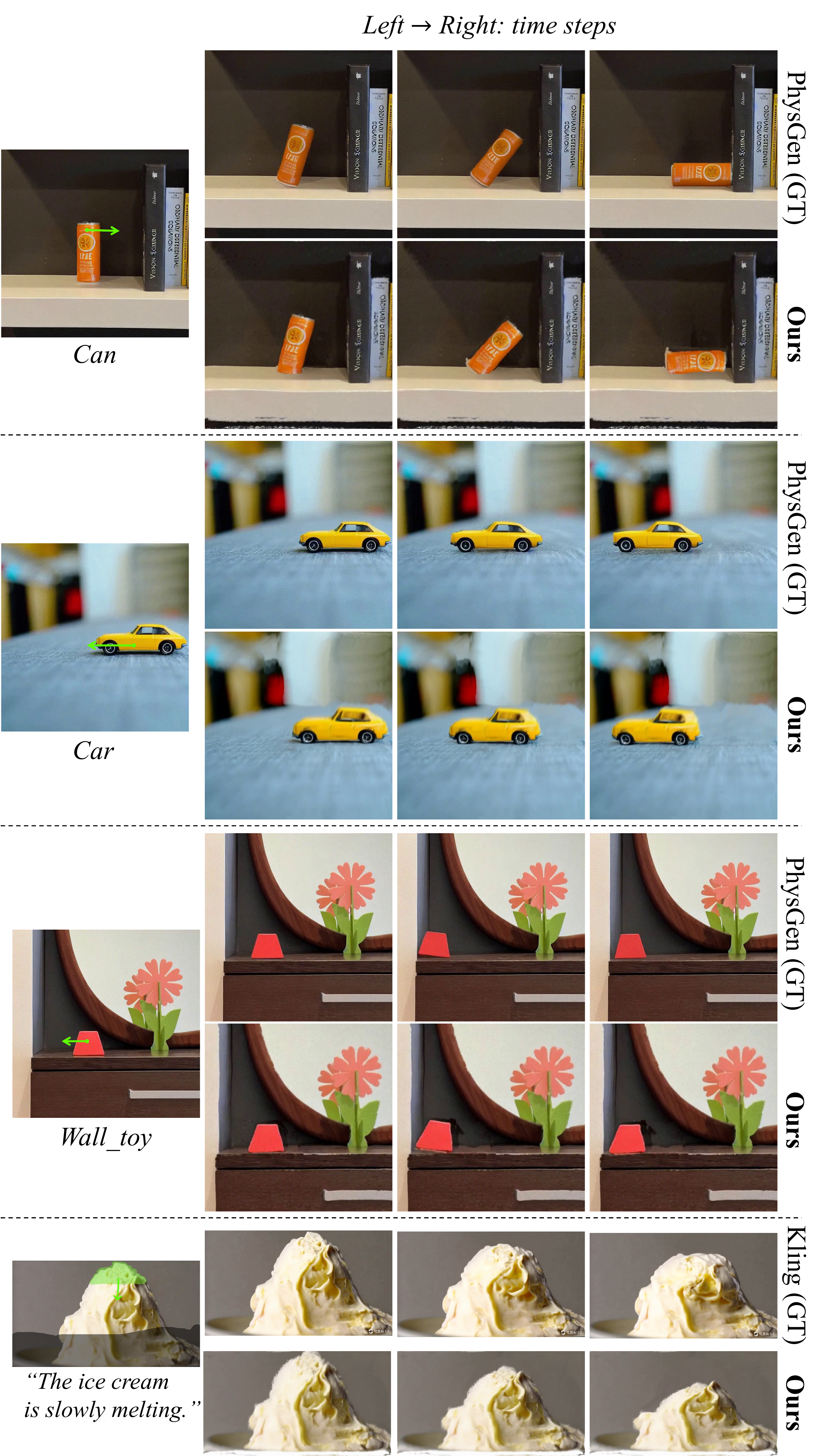}
    \vspace{-0.5cm}
    \caption{Qualitative results of our method using video guidance from PhysGen and Kling. The green arrows show the input force for the simulated objects. For Kling video generation, we utilize a text prompt and an input image, complemented by a motion brush (\textit{green mask}) to define the motion trajectory (\textit{green arrow}) and a static mask (\textit{gray mask}) to restrict camera movement.}
    \label{fig:supp_others}
\end{figure}

To evaluate our approach, we perform experiments using videos generated by PhysGen and Kling~\cite{kling}, with input images sourced from the PhyGenBench~\cite{phygenbench} dataset. As shown in Fig.~\ref{fig:supp_others}, the visual results highlight the capability of our method to capture realistic deformations and motion trajectories guided by videos from both PhysGen and Kling. We also report the quantitative results (ECMS) on PhysGen scenes in Tab.~\ref{tab:comp_physgen}. These results demonstrate the adaptability and robustness of our method, even when limited to single-view inputs, effectively simulating complex interactions and maintaining high fidelity across varying scenarios.

\subsection{Ablation Study}\label{sec:supp_abla}
\begin{figure}[ht]
    \centering
    \includegraphics[width=1.0\linewidth]{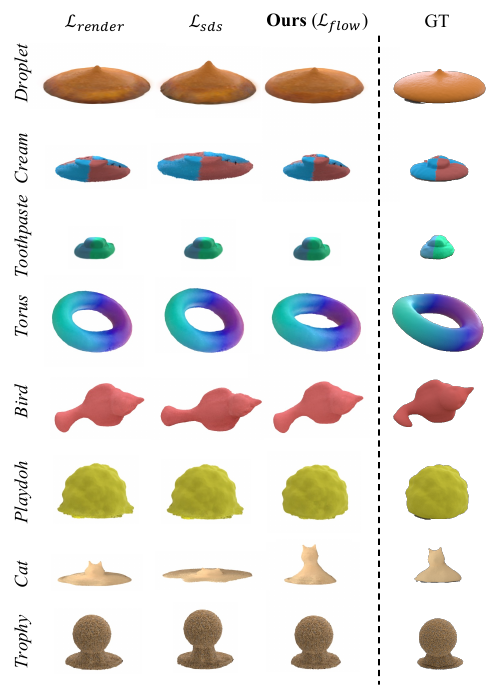}
    \vspace{-0.6cm}
    \caption{Qualitative results of ablation study comparing different loss functions for system identification on synthetic dataset.}
    \label{fig:supp_abla_syn}
\end{figure}

\noindent\textbf{Effectiveness of Optical Flow Guidance:}
We performed ablations of $\mathcal{L}_{flow}$ on the entire synthetic dataset, which includes 9 cases across various material types. We provide detailed ablation results of each object in Tab.~\ref{tab:supp_abla_syn} and qualitative comparisons in Fig.~\ref{fig:supp_abla_syn}. Our method consistently achieves the lowest relative error (RE) and ECMS across a range of material parameters compared to other loss functions. The simulations generated by our approach closely align with the ground-truth, both in terms of material deformation and shape fidelity. These results show that optical flow guidance effectively captures complex material behaviors and enables precise material parameter optimization.

\begin{figure}[t]
  \centering
   \includegraphics[width=\linewidth]{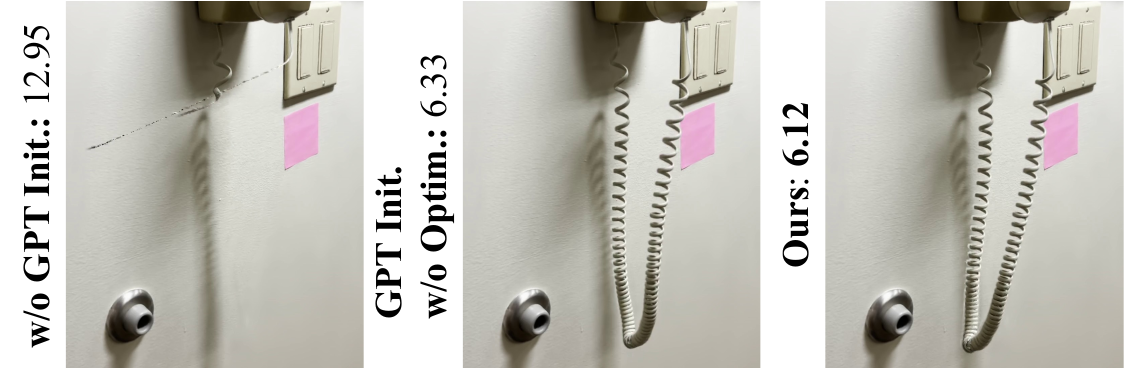}
   \vspace{-0.7cm}
   \caption{Ablations of GPT initialization with ECMS$\downarrow$.}
   \label{fig:abla_gpt}
    \vspace{-0.3cm}
\end{figure}

\noindent\textbf{Effectiveness of GPT Initialization:}
Along with experimental results in Sec.~\ref{sec:abla_gpt} and Fig.~\ref{fig:abla_gpt}, GPT initialization reduces errors, and our full method with optical flow guidance achieves the lowest ECMS score and realistic motion.

% -------------------------
\section{Limitations}
Our method simulates deformations on 3D Gaussian splats and renders the resulting frames without incorporating relighting effects. This limits visual realism in aspects such as dynamic shadows and specular highlights. Future work could focus on integrating relighting techniques to capture more complex lighting interactions, thereby enhancing the overall fidelity and realism of the simulated scenes.

\clearpage

\clearpage

\begin{table*}[!t]\tabcolsep=0.6cm
    \centering
    \resizebox{1.\linewidth}{!}{
    \begin{tabular}{llll}
    \toprule
    \textbf{Object} & $\mathcal{L}_{render}$ & $\mathcal{L}_{sds}$ & \textbf{Ours} ($\mathcal{L}_{flow}$) \\ \hline

    Droplet & $\delta_\mu=0.045$, $\delta_\kappa=0.080$ & $\delta_\mu=\mathbf{0.005}$, $\delta_\kappa=0.820$ & $\delta_\mu=0.004$, $\delta_\kappa=\mathbf{0.030}$ \\ 
    Letter & $\delta_\mu=0.162$, $\delta_\kappa=0.350$ & $\delta_\mu=0.050$, $\delta_\kappa=\mathbf{0.000}$ & $\delta_\mu=\mathbf{0.023}$, $\delta_\kappa=0.893$\\ \hline
    % -----------
    \multirow{2}{*}{Cream} & $\delta_\mu=11.100$, $\delta_\kappa=0.570$, & $\delta_\mu=\mathbf{0.030}$, $\delta_\kappa=\mathbf{0.480}$, & $\delta_\mu=\mathbf{0.080}$, $\delta_\kappa=0.755$, \\
    & $\delta_{\tau_Y}=0.053$, $\delta_\eta=0.440$ & $\delta_{\tau_Y}=\mathbf{0.007}$, $\delta_\eta=\mathbf{0.340}$ & $\delta_{\tau_Y}=0.577$, $\delta_\eta=0.750$ \\
    \multirow{2}{*}{Toothpaste} & $\delta_\mu=0.302$, $\delta_\kappa=1.220$, & $\delta_\mu=0.162$, $\delta_\kappa=\mathbf{0.076}$, & $\delta_\mu=\mathbf{0.015}$, $\delta_\kappa=0.136$, \\
    & $\delta_{\tau_Y}=0.140$, $\delta_\eta=\mathbf{0.023}$ & $\delta_{\tau_Y}=\mathbf{0.130}$, $\delta_\eta=0.090$ & $\delta_{\tau_Y}=0.232$, $\delta_\eta=0.539$ \\ \hline
    % -----------
    Torus & $\delta_E=0.040$, $\delta_\nu=0.073$ & $\delta_E=\mathbf{0.010}$, $\delta_\nu=\mathbf{0.017}$ & $\delta_E=0.039$, $\delta_\nu=0.989$ \\ 
    Bird & $\delta_E=0.073$, $\delta_\nu=0.090$ & $\delta_E=\mathbf{0.027}$, $\delta_\nu=\mathbf{0.053}$ & $\delta_E=0.040$, $\delta_\nu=0.571$ \\ \hline
    % -----------
    \multirow{2}{*}{Playdoh} & $\delta_E=0.920$, $\delta_\nu=0.093$, & $\delta_E=0.210$, $\delta_\nu=\mathbf{0.073}$, & $\delta_E=\mathbf{0.104}$, $\delta_\nu=0.327$, \\
    & $\delta_{\tau_Y}=0.097$ & $\delta_{\tau_Y}=\mathbf{0.013}$ & $\delta_{\tau_Y}=0.027$ \\ 
    \multirow{2}{*}{Cat} & $\delta_E=0.839$, $\delta_\nu=0.023$, & $\delta_E=\mathbf{0.020}$, $\delta_\nu=\mathbf{0.013}$, & $\delta_E=0.387$, $\delta_\nu=0.414$,  \\
    & $\delta_{\tau_Y}=0.073$ & $\delta_{\tau_Y}=\mathbf{0.023}$ & $\delta_{\tau_Y}=0.221$ \\ \hline
    % -----------
    Trophy & $\delta_{\theta_{fric}}=0.098$ & $\delta_{\theta_{fric}}=0.117$ & $\delta_{\theta_{fric}}=\mathbf{0.013}$ \\ 
    \bottomrule
    \end{tabular}
    }
    \vspace{-0.2cm}
    \caption{Comparisons with baselines for system identification performance on the synthetic dataset. $\delta_{*}$ denotes the relative error (RE) $\downarrow$ for the material parameter $*$.}
    \label{tab:supp_comp_syn}
\end{table*}

\begin{table*}[!t]\tabcolsep=0.6cm
    \centering
    \resizebox{1.\linewidth}{!}{
    \begin{tabular}{llll}
    \toprule
    \textbf{Object} & $\mathcal{L}_{render}$ & $\mathcal{L}_{sds}$ & \textbf{Ours} ($\mathcal{L}_{flow}$) \\ \hline

    Droplet & $\delta_\mu=0.230$, $\delta_\kappa=0.731$ & $\delta_\mu=1.515$, $\delta_\kappa=0.449$ & $\delta_\mu=\mathbf{0.004}$, $\delta_\kappa=\mathbf{0.030}$ \\ 
    Letter & $\delta_\mu=0.250$, $\delta_\kappa=0.918$ & $\delta_\mu=0.575$, $\delta_\kappa=\mathbf{0.827}$ & $\delta_\mu=\mathbf{0.023}$, $\delta_\kappa=0.893$\\ \hline
    % -----------
    \multirow{2}{*}{Cream} & $\delta_\mu=0.160$, $\delta_\kappa=0.732$, & $\delta_\mu=3.320$, $\delta_\kappa=\mathbf{0.004}$, & $\delta_\mu=\mathbf{0.080}$, $\delta_\kappa=0.755$, \\
    & $\delta_{\tau_Y}=\mathbf{0.070}$, $\delta_\eta=0.841$ & $\delta_{\tau_Y}=0.843$, $\delta_\eta=0.893$ & $\delta_{\tau_Y}=0.577$, $\delta_\eta=\mathbf{0.750}$ \\
    \multirow{2}{*}{Toothpaste} & $\delta_\mu=0.283$, $\delta_\kappa=0.173$, & $\delta_\mu=0.255$, $\delta_\kappa=0.141$, & $\delta_\mu=\mathbf{0.015}$, $\delta_\kappa=\mathbf{0.136}$, \\
    & $\delta_{\tau_Y}=0.480$, $\delta_\eta=0.249$ & $\delta_{\tau_Y}=0.375$, $\delta_\eta=\mathbf{0.178}$ & $\delta_{\tau_Y}=\mathbf{0.232}$, $\delta_\eta=0.539$ \\ \hline
    % -----------
    Torus & $\delta_E=0.237$, $\delta_\nu=\mathbf{0.586}$ & $\delta_E=0.369$, $\delta_\nu=0.771$ & $\delta_E=\mathbf{0.039}$, $\delta_\nu=0.989$ \\ 
    Bird & $\delta_E=0.120$, $\delta_\nu=0.616$ & $\delta_E=0.797$, $\delta_\nu=0.727$ & $\delta_E=\mathbf{0.040}$, $\delta_\nu=\mathbf{0.571}$ \\ \hline
    % -----------
    \multirow{2}{*}{Playdoh} & $\delta_E=0.483$, $\delta_\nu=0.196$, & $\delta_E=0.828$, $\delta_\nu=\mathbf{0.165}$, & $\delta_E=\mathbf{0.104}$, $\delta_\nu=0.327$, \\
    & $\delta_{\tau_Y}=0.942$ & $\delta_{\tau_Y}=0.943$ & $\delta_{\tau_Y}=\mathbf{0.027}$ \\ 
    \multirow{2}{*}{Cat} & $\delta_E=\mathbf{0.167}$, $\delta_\nu=\mathbf{0.290}$, & $\delta_E=0.644$, $\delta_\nu=0.623$, & $\delta_E=0.387$, $\delta_\nu=0.414$,  \\
    & $\delta_{\tau_Y}=0.652$ & $\delta_{\tau_Y}=1.288$ & $\delta_{\tau_Y}=\mathbf{0.221}$ \\ \hline
    % -----------
    Trophy & $\delta_{\theta_{fric}}=0.173$ & $\delta_{\theta_{fric}}=0.305$ & $\delta_{\theta_{fric}}=\mathbf{0.013}$ \\ 
    \bottomrule
    \end{tabular}
    }
    \vspace{-0.2cm}
    \caption{Ablation study of different losses for system identification performance on the synthetic dataset. $\delta_{*}$ denotes the relative error (RE) $\downarrow$ for the material parameter $*$.}
    \label{tab:supp_abla_syn}
\end{table*}

% {
%     \small
%     \bibliographystyle{ieeenat_fullname}
%     \bibliography{main}
% }

\end{document}